\documentclass[11pt]{article}

\usepackage[final]{acl}

\usepackage{times}
\usepackage{latexsym}

\usepackage[T1]{fontenc}

\usepackage[utf8]{inputenc}

\usepackage{microtype}

\usepackage{inconsolata}

\usepackage{graphicx}

\newcommand{\scare}[1]{`#1'}

\title{NLG Evaluation: Past, Present, Future}

\author{Ehud Reiter \\
  Dept of Computing Science \\
  University of Aberdeen\\
  Aberdeen, UK \\
  \texttt{e.reiter@abdn.ac.uk} \\}

\begin{document}
\maketitle
\begin{abstract}
Natural Language Generation (NLG) evaluation has changed dramatically since 1990, and will continue to evolve in the future.  In 1990, when NLG had close ties to linguistics, there was very little formal experimental evaluation in the modern sense.  In 2026, when NLG is closely linked to machine learning,  experimental evaluation is expected and indeed fundamental to research.  Many evaluation techniques were developed over this period, including most recently LLM-as-Judge.   I expect NLG evaluation will continue to evolve in the future.  In particular, impact, qualitative, and safety evaluation will become more important as large numbers of people routinely use NLG technology.
\end{abstract}

\section{Introduction}

The evaluation of Natural Language Generation (NLG) systems has changed dramatically over my career.  In 1990, when I got my PhD in NLG, most NLG research papers did not include a quantitative experimental evaluation of a research question.
%Instead, they presented ideas and supporting evidence, in a way which was sometimes similar to linguistics papers.
By 2026, NLG research papers are expected to include structured experimental evaluations of hypotheses, although the quality and validity of these evaluations is variable.  I expect that by 2036, impact, safety, and qualitative evaluations will be much more important, because  NLG technology will be widely used by large numbers of people.  Table~\ref{tab:evalHistory} summarises my view of NLG evaluation at different points in time.

\section{NLG Evaluation in the Past}

\subsection{1990: Little quantitative experimental evaluation}
The International NLG (INLG) conference in 1990 had 25 papers.  \textit{None} of them included a structured quantitative hypothesis test.  Instead, these papers mostly presented an algorithm, technique, resource, or system, and justified it on engineering or linguistic criteria.  For example, \citet{mccoy-etal-1990-using} proposed combining tree-adjoining and systemic grammars, and justified this by arguing that their approach did a better job of handling long-distance dependencies (linguistics) and also makes it easier to build grammars (engineering).  Their argument was qualitative, no numbers were given.

In the broader NLP world, the speech recognition community had adapted the idea of quantitative comparisons of the performance of systems \citep{waibel1990readings}, but this was unusual in the rest of NLP.  Perhaps the most important NLP paper in 1990 was \citet{brown-etal-1990-statistical}, which introduced statistical machine translation, but even it did not provide quantitative comparisons of the sort we expect in 2026.

\begin{table*}
    \centering
    \begin{tabular}{|l|p{6cm}|p{7cm}|} \hline
        \textbf{year}  & \textbf{NLG evaluation} & \textbf{example paper and its evaluation} \\ \hline
        1990  & non-quantitative evaluation, often using linguistic or engineering arguments & \citet{mccoy-etal-1990-using}: qualitative argument that their grammatical approach handles long-distance dependencies better\\ \hline
        2000 & wide mix of different techniques, including metrics, human ratings, and task-based & \citet{cheng-mellish-2000-empirical}: use human ratings to evaluate different ways of expressing causal and temporal relationships \\ \hline
        2010 & standardised evaluation techniques and shared tasks based on these & \citet{belz-kow-2010-grec}: results of GREC shared task on generating referring expressions \\ \hline
        2020 & research on evaluation becomes an important research area & \citet{howcroft-etal-2020-twenty}: gives recommendations for reporting human evaluations, based on meta-analysis of published evaluations\\ \hline
        2026 & LLM-as-Judge, annotation by human experts, safety, interdisciplinary &\citet{bean2026reliability}: use medical evaluation techniques to assess system that answers health queries \\ \hline
        2036 & impact, qualitative, safety evaluation & \textit{not yet written} \\ \hline
    \end{tabular}
    \caption{NLG evaluation over the years}
    \label{tab:evalHistory}
\end{table*}

\subsection{2000: Wide range of evaluation techniques}
INLG in 2000 had 38 papers, and these included many different kinds of evaluation, as well as one of the first paper that was \textit{about} evaluation \cite{bangalore-etal-2000-evaluation}.  Types of evaluation included
\begin{itemize}
    \item Human evaluation \citep{cheng-mellish-2000-empirical}
    \item Metric-based evaluation \citep{minnen-etal-2000-robust}
    \item Task-based evaluation \citep{carenini-2000-task}
\end{itemize}
There were also papers which continued to assess their contribution using engineering or linguistic arguments, as in 1990.

In short, by 2000 experimental evaluations was recognised as being important.  However there were no widely accepted standard evaluation techniques in NLG. 

A similar mix was seen at larger NLP events such as ACL.  Evaluation was clearly regarded as important, but many techniques were being tried.  The broader NLP community focused more on metric-based evaluation, including \citet{gildea-jurafsky-2000-automatic}, which won a Test of Time award.  However ACL in this period also included papers reporting complex task-based evaluations \cite{mani-etal-1999-tipster,reiter-etal-2001-using}.

\subsection{2010: Shared tasks and standard evaluations}
INLG in 2010 had 37 papers, many of which were shared task submissions.  Shared tasks (such as the GREC challenge for generating referring expressions \citep{belz-kow-2010-grec}) had become an accepted part of NLG as well as NLP research, and used metrics and/or human evaluations to evaluate the performance of submissions.
%Some papers did evaluations in simulated environments \citep{dethlefs-cuayahuitl-2010-hierarchical}.   There also continued to be papers which used task-based evaluations and linguistic assessments.
Some papers also began to describe evaluations in considerable detail \citep{murray-etal-2010-generating}.

The wider NLP community had embraced ngram-based metrics for evaluation of text production, and the BLEU and ROUGE metrics had effectively become standards.  Papers in machine translation were expected to use BLEU, and papers in summarisation were expected to use ROUGE.

Human evaluation had become unusual in ACL conferences, although the annual WMT shared task continued to use it.  The NLG community, however, insisted on using human evaluations, and could point to papers which suggested that metrics were not reliable in NLG \citep{reiter-belz-2009-investigation}.  When doing human evaluations, most researchers either used Likert scales or asked subjects to rank a set of texts by a quality criteria; these became standard techniques for human evaluation of generated texts.

\subsection{2020: Evaluation is important research area}
INLG in 2020 had 46 papers (it has not seen the exponential growth that ACL has had in recent years).   Perhaps the most notable change compared to 2010 was that evaluation has become a very important part of the community's research agenda.  Indeed both of the INLG2020 best papers were about evaluation \citep{belz-etal-2020-disentangling,dusek-kasner-2020-evaluating}, and there were several other papers about evaluation methodology \citep{howcroft-etal-2020-twenty,thomson-reiter-2020-gold} in INLG2020.

The wider NLP community also placed increasing importance on evaluation as an important research theme.  For example the ACL 2020 best paper was about testing \citep{ribeiro-etal-2020-beyond}, and one of the two honourable mention papers was about evaluation \citep{mathur-etal-2020-tangled}.

In short, evaluation was now not just something which researchers had to do, but also an important research topic in its own right.

\section{NLG Evaluation in 2026}

%In 2026, NLG and NLP (and indeed society at large) are changing because of large language models (LLMs).  The fields have exploded in size, and commercial companies are investing huge amounts of money.  The distinction between NLG and other areas of NLP is fading, since NLG and NLP researchers both use LLM technology.

\citet{reiter2025} summarised NLG evaluation in 2025, including links to papers that gave best practice suggestions.  Large language model (LLM) technology had become widespread and this had changed NLG evaluation and introduced new challenges.

\subsection{Evaluation challenges from LLMs}
There are many challenges in evaluating LLMs, including the following.

\textit{Higher quality generated texts:}  Texts produced by LLMs are usually higher quality than texts produced by previous technologies (rule-based, LSTM), and can in some cases be human quality, or even better-than-human.  This means that many traditional evaluation techniques, such as metrics that compare generated texts against human-written reference texts, no longer work well.  If we expect a generated text to be better than human, then evaluating it by comparing it to a human-written reference text does not make sense.

\textit{Semantic and pragmatic evaluations:}  Texts produced by LLMs are almost always fluent and readable, so evaluating readability is less useful.  Instead, there is more emphasis on evaluating semantic and pragmatic quality criteria \citep{reiter2025}, such as accuracy/hallucinations, omissions, and contextual appropriateness.

\textit{Data contamination:}  Since LLMs are trained on the Internet, an evaluation that uses Internet data may not mean much, since the LLM may have memorised the test data \citep{balloccu2024leak}. 

\textit{Worst-case and safety evaluation:} The growing real-world usage of LLMs in safety-critical contexts such as medicine (where flawed texts could harm patients) means that we need techniques that evaluate \scare{worst-case} performance of LLMs \citep{reiter2025}.  If a medical LLM gives good output in 99.9\% of cases but harmful output in 0.1\% of cases, this is not acceptable. 

\textit{Interdisciplinary interest and usage:} The growing real-world usage of LLMs means that other disciplines (such as medicine and law) want LLM-based NLG systems to be evaluated using their methodologies and expectations \citep{duggan:Jama}.

\subsection{Changes in NLG evaluation}
The above challenges have changed the way NLG is evaluated.

\textit{LLM as Judge:}  The above problems have stimulated interest in reference-free metrics which work for semantic and pragmatic quality criteria, including in particular using LLMs to evaluate the quality of texts produced by other LLMs \citep{gao-etal-2025-llm}; this is called \textit{LLM as Judge}.  This seems to work well in some cases but not others; unfortunately many researchers use LLM evaluators without checking that they are effective in their use case.

\textit{Human evaluation using expert annotations:} Human evaluations in NLG have traditionally used Likert-type rating scales.  This seems to work less well when evaluating semantic and pragmatic problems in high-quality LLM texts, especially with crowdworkers (who may cheat by using LLMs to do the evaluation task \cite{asher2026}). We are seeing more human evaluations that instead ask knowledgable people to annotate specific problems in a generated text \citep{THOMSON2023101482}.

\textit{Private test data:}  In 2020, test data sets were typically published (e.g., on GitHub repos), which made replication easier.  But in 2026, data contamination concerns mean  that test data is sometimes not published or shared.

\textit{Safety evaluations:} Many techniques have been proposed for safety evaluation.  This area is heavily influenced by cyber security, and includes techniques for risk analysis (such as red teaming), risk mitigation (e.g., monitoring), and risk governance (such as incident reporting)
\citep{bengio2026internationalaisafetyreport}.

\textit{Interdisciplinary evaluations:} High-quality evaluations of NLG systems are appearing in other fields, notably medicine, that use medical evaluation techniques such as randomised controlled trials (which are very rare in the NLP literature \citep{reiter-2025-evaluate}).  Sometimes these give different results from classical NLP evaluation, which raises important questions about the best way to evaluate NLG .

\subsection{Ongoing challenges for NLG evaluation}
The new evaluation techniques described above are being adopted by many researchers and help in addressing some of the new evaluation challenges of LLMs.  But there are many  problems and concerns that still need to be addressed.  These include
\begin{itemize}
    \item \textit{Experimental rigour:} Unfortunately, many experiments are poorly designed, poorly executed, or distorted by bugs \citep{thomson-etal-2024-common}.
    \item \textit{Replicability:} Many experiments cannot be replicated, in part because their authors do not support replication \citep{belz-etal-2023-non}.
    \item \textit{Construct validity:} Many evaluation techniques, especially benchmarks, do not measure what they claim to measure \citep{bean2025measuring}.
    \item \textit{Cheating:}  LLMs engage in behaviour such as reward hacking \cite{recent-frontier-models-are-reward-hacking}, which is essentially cheating. \citet{asadi2026mirageillusionvisualunderstanding} show that LLMs can get very high benchmark scores even when input data is withheld, by picking up on subtle clues in the wording of questions in the benchmark.
    \item \textit{Commercial bias and incentives:} A lot of evaluation research and development is funded by AI companies such as OpenAI, who have an interest in ensuring that their systems do well on these evaluations \citep{cheng2025benchmarking}.
    \item \textit{Evolving benchmarks:} New evaluation benchmarks are constantly being proposed, and existing benchmarks often  become saturated \citep{akhtar2026aibenchmarksplateausystematic} and hence useless.  It is difficult for many researchers to stay up-to-date on the best benchmark to use.
\end{itemize}

A generic challenge is that the research culture in NLP is often not very supportive of high quality evaluation.   Many people feel pressure to publish large numbers of papers, and reviewers often show limited interest in quality of data sets, validity of evaluation metrics, experimental rigour, etc.  This encourages researchers to conduct \scare{quick and dirty} evaluations.

%In the past, one could argue that while evaluations in medicine impacted the real world, and poor evaluations could hurt people, evaluations in NLP and AI had limited real-world impact, so it was less important if they were flawed.  But the advent of LLMs means that AI and NLP may be used by billions of people, so poor evaluation can hurt people.

\section{NLG Evaluations in the Future}
What will NLG evaluation be like in ten years time (2036)?  The above challenges will hopefully be addressed, but more generally we also need to go beyond measuring performance on a test set, which dominates NLG and NLP evaluation in 2026.  If we care about how our technology affects the real world, we need to do more of the following:
\begin{itemize}
\item Directly measure the real-world \textbf{impact} of NLG systems.
\item Use \textbf{qualitative} techniques to get insights about the effectiveness of our techniques in messy and complex real-world contexts.
\item Analyse what happens in worst-case or adversarial contexts, especially for \textbf{safety} criteria.
\end{itemize}
These techniques will help ensure that evaluation is relevant and meaningful in a future world where NLG is a widely-used technology. 

Note that impact, qualitative, and safety evaluations are not new, they are already being done in 2026 to a limited degree in NLG; they are much more common in Medicine, perhaps because medical research has had real-world consequences for decades or indeed centuries.  So the challenge for the NLG community is to embrace these types of evaluation and learn how to do them well in an NLG context.

The spread of more types of NLG evaluation may lead to the evolution of evaluation frameworks, which show how different types of evaluation can be combined to obtain a holistic understanding of what a system can do \cite{reddy2021evaluation}.

\subsection{Impact evaluation}
As discussed by \citet{reiter2025}, there is very little evaluation of real-world impact in the NLP and NLG research literature, by which we mean how real-world usage of an NLG system changes key performance indicators (KPIs).  As NLG technology improves and becomes more widely used, we need more impact studies, especially if we want to measure utility in messy real-world contexts.
%The utility and usefulness of an NLG system depends on many factors, including context, workflow, and user training, so it is difficult to predict from metrics or human evaluations in artificial contexts.

A good example is \citet{bean2026reliability}, which measured how well LLMs can respond to health queries based on scenarios.  LLMs do well at this task if given the scenario directly, or if they interact with an LLM-simulated user.  However, if they interact with human users (who often communicate in a confused way), their performance is much worse.  Hence if we want to genuinely evaluate how well an LLM can respond to health queries, we need to measure what happens when real people interact with the LLM.
Ideally, this should be based on real patients asking about their health problems \citep{brodeur2026prospectiveclinicalfeasibilitystudy}. 

There are many ways to evaluate impact, including randomised controlled trials (RCT), A/B tests, before-and-after (pre-post) studies, and observational studies \citep{reiter-2025-evaluate}.
By 2036, we hope that such evaluations will much more common.  Most NLG evaluations will probably still use simpler and cheaper techniques, but a significant number will evaluate real-world impact.

\subsection{Qualitative evaluation}
Evaluation in NLG and NLP is almost always quantitative, and typically uses statistical hypothesis testing.  Such evaluation is very important, but should be supplemented by qualitative evaluation, which can provide additional insights which are very useful in complex real-world contexts \citep{greenhaigh1997qualitative,tisdell2025qualitative}.

Some qualitative evaluation techniques are already used in NLG, including error analysis \citep{van-miltenburg-etal-2023-barriers} and analysis of free-text comments from participants \citep{VANDERLEE2021101151}.  But many other techniques are rarer, including data collection techniques such as focus groups \citep{Sun2026.02.02.26345346} and (semi-)structured interviews \citep{zhou-etal-2022-deconstructing}, and analysis techniques such as thematic analysis \cite{guest2011applied} and
content analysis \citep{sambaraju-etal-2011-text}.

As NLG systems become more capable and are used in a wider variety of complex contexts, we expect that qualitative evaluation and insights will become more important, especially since many quantitative results will quickly become dated as newer models are released.
%By 2036, we expect that qualitative evaluation will be much more common and indeed expected.

\subsection{Safety evaluation}
Safety evaluation is not new, it is a rapidly growing area of evaluation, which looks at whether AI systems can harm individuals (for example by encouraging suicide\footnote{See \mbox{\url{https://www.thehumanlineproject.org/}} stories, such as %\url{https://www.theguardian.com/society/2026/mar/31/teenager-asked-chatgpt-most-successful-ways-take-life-inquest-told}
\citet{guardian2026}} or giving dangerous medical advice \citep{bickmore2018patient}) or society (e.g., by empowering hackers or terrorists) \citep{bengio2026internationalaisafetyreport}.

We expect that safety will become one of the main foci of evaluation research.  Ultimately, performance evaluation is of interest primarily to companies and academics who develop NLG technology, whereas safety evaluation is of interest to everyone who \textit{uses} NLG technology, which is a much larger number of people.  Safety evaluation is probably more important to society than performance evaluation.  Indeed, governments have begun to impose safety standards on AI systems\footnote{\url{https://www.gov.uk/government/publications/generative-ai-product-safety-standards}}, and this may lead to formal government involvement in AI evaluation methodology.

Safety evaluation is also more challenging than performance evaluation, because it is about worst-case behaviour, and behaviour under adversarial attack (e.g., hackers trying to break into a system).  Performance evaluations usually look at average case performance, so they can be computed based on a representative sample.  Safety evaluation requires looking for misbehaviour everywhere, including edge cases, which are hard to predict for complex stochastic black box neural models.  It will almost certainly require monitoring of the actual behaviour of deployed systems, as well as experiments on test data or test subjects.

\section{Conclusion}
NLG evaluation has changed dramatically between 1990 (mostly lingiustic evaluation) and 2026 (LLM-as-Judge and human annotation protocols).  It continues to evolve, and the next ten years should be exciting, with more focus on impact, qualitative, and safety evaluation.

\section*{Acknowledgements}
Many thanks to the anonymous reviewers, the members of the Aberdeen CLAN research group, and Saad Mahamood for their very helpful comments.

\bibliography{nlgeval}

@misc{akhtar2026aibenchmarksplateausystematic,
      title={When AI Benchmarks Plateau: A Systematic Study of Benchmark Saturation}, 
      author={Mubashara Akhtar and Anka Reuel and Prajna Soni and Sanchit Ahuja and Pawan Sasanka Ammanamanchi and Ruchit Rawal and Vilém Zouhar and Srishti Yadav and Chenxi Whitehouse and Dayeon Ki and Jennifer Mickel and Leshem Choshen and Marek Šuppa and Jan Batzner and Jenny Chim and Jeba Sania and Yanan Long and Hossein A. Rahmani and Christina Knight and Yiyang Nan and Jyoutir Raj and Yu Fan and Shubham Singh and Subramanyam Sahoo and Eliya Habba and Usman Gohar and Siddhesh Pawar and Robert Scholz and Arjun Subramonian and Jingwei Ni and Mykel Kochenderfer and Sanmi Koyejo and Mrinmaya Sachan and Stella Biderman and Zeerak Talat and Avijit Ghosh and Irene Solaiman},
      year={2026},
      eprint={2602.16763},
      archivePrefix={arXiv},
      primaryClass={cs.AI},
      url={https://arxiv.org/abs/2602.16763}, 
}

@misc{asadi2026mirageillusionvisualunderstanding,
      title={MIRAGE: The Illusion of Visual Understanding}, 
      author={Mohammad Asadi and Jack W. O'Sullivan and Fang Cao and Tahoura Nedaee and Kamyar Rajabalifardi and Fei-Fei Li and Ehsan Adeli and Euan Ashley},
      year={2026},
      eprint={2603.21687},
      archivePrefix={arXiv},
      primaryClass={cs.AI},
      url={https://arxiv.org/abs/2603.21687}, 
}

@article{asher2026,
author = {Michael W. Asher and Gillian Gold and Eason Chen and Paulo F. Carvalho},
title ={Chatbots Are Undermining Crowdsourced Research in the Behavioral Sciences: Detecting Artificial Intelligence–Assisted Cheating With a Keystroke-Based Tool},
journal = {Advances in Methods and Practices in Psychological Science},
volume = {9},
number = {1},
pages = {25152459261424723},
year = {2026},
doi = {10.1177/25152459261424723},
URL = { 
        https://doi.org/10.1177/25152459261424723
},
eprint = { 
        https://doi.org/10.1177/25152459261424723}
}

@article{guardian2026,
  author       = {Nadeem Badshah},
  title        = {Teenager died after asking ChatGPT for ‘most successful’ way to take his life, inquest told},
  journal      = {The Guardian},
  year         = {2026},
  month        = {March},
  day          = {31},
  url          = {https://www.theguardian.com/society/2026/mar/31/teenager-asked-chatgpt-most-successful-ways-take-life-inquest-told}
}

@inproceedings{balloccu2024leak,
  title={Leak, cheat, repeat: Data contamination and evaluation malpractices in closed-source LLMs},
  author={Balloccu, Simone and Schmidtov{\'a}, Patr{\'\i}cia and Lango, Mateusz and Du{\v{s}}ek, Ond{\v{r}}ej},
  booktitle={Proceedings of the 18th Conference of the European Chapter of the Association for Computational Linguistics (Volume 1: Long Papers)},
  pages={67--93},
  year={2024}
}

@inproceedings{bangalore-etal-2000-evaluation,
    title = "Evaluation Metrics for Generation",
    author = "Bangalore, Srinivas  and
      Rambow, Owen  and
      Whittaker, Steve",
    editor = "Elhadad, Michael",
    booktitle = "{INLG}{'}2000 Proceedings of the First International Conference on Natural Language Generation",
    month = jun,
    year = "2000",
    address = "Mitzpe Ramon, Israel",
    publisher = "Association for Computational Linguistics",
    url = "https://aclanthology.org/W00-1401/",
    doi = "10.3115/1118253.1118255",
    pages = "1--8"
}

@article{bean2025measuring,
  title={Measuring what matters: Construct validity in large language model benchmarks},
  author={Bean, Andrew M and Kearns, Ryan Othniel and Romanou, Angelika and Hafner, Franziska Sofia and Mayne, Harry and Batzner, Jan and Foroutan, Negar and Schmitz, Chris and Korgul, Karolina and Batra, Hunar and others},
  journal={arXiv preprint arXiv:2511.04703; presented at Neurips 2025},
  year={2025}
}

@article{bean2026reliability,
  title={Reliability of LLMs as medical assistants for the general public: a randomized preregistered study},
  author={Bean, Andrew M and Payne, Rebecca Elizabeth and Parsons, Guy and Kirk, Hannah Rose and Ciro, Juan and Mosquera-G{\'o}mez, Rafael and Hincapi{\'e} M, Sara and Ekanayaka, Aruna S and Tarassenko, Lionel and Rocher, Luc and others},
  journal={Nature Medicine},
  pages={1--7},
  year={2026},
  publisher={Nature Publishing Group US New York}
}

@inproceedings{belz-kow-2010-grec,
    title = "The {GREC} Challenges 2010: Overview and Evaluation Results",
    author = "Belz, Anja  and
      Kow, Eric",
    editor = "Kelleher, John  and
      Namee, Brian Mac  and
      Sluis, Ielka van der",
    booktitle = "Proceedings of the 6th International Natural Language Generation Conference",
    month = jul,
    year = "2010",
    publisher = "Association for Computational Linguistics",
    url = "https://aclanthology.org/W10-4226/"
}

@inproceedings{belz-etal-2020-disentangling,
    title = "Disentangling the Properties of Human Evaluation Methods: A Classification System to Support Comparability, Meta-Evaluation and Reproducibility Testing",
    author = "Belz, Anya  and
      Mille, Simon  and
      Howcroft, David M.",
    editor = "Davis, Brian  and
      Graham, Yvette  and
      Kelleher, John  and
      Sripada, Yaji",
    booktitle = "Proceedings of the 13th International Conference on Natural Language Generation",
    month = dec,
    year = "2020",
    address = "Dublin, Ireland",
    publisher = "Association for Computational Linguistics",
    url = "https://aclanthology.org/2020.inlg-1.24/",
    doi = "10.18653/v1/2020.inlg-1.24",
    pages = "183--194",
}

@inproceedings{belz-etal-2023-non,
    title = "Non-Repeatable Experiments and Non-Reproducible Results: The Reproducibility Crisis in Human Evaluation in {NLP}",
    author = "Belz, Anya  and
      Thomson, Craig  and
      Reiter, Ehud  and
      Mille, Simon",
    editor = "Rogers, Anna  and
      Boyd-Graber, Jordan  and
      Okazaki, Naoaki",
    booktitle = "Findings of the Association for Computational Linguistics: ACL 2023",
    month = jul,
    year = "2023",
    address = "Toronto, Canada",
    publisher = "Association for Computational Linguistics",
    url = "https://aclanthology.org/2023.findings-acl.226/",
    doi = "10.18653/v1/2023.findings-acl.226",
    pages = "3676--3687",
}

@misc{bengio2026internationalaisafetyreport,
      title={International AI Safety Report 2026}, 
      author={Yoshua Bengio and Stephen Clare and Carina Prunkl and Maksym Andriushchenko and Ben Bucknall and Malcolm Murray and Rishi Bommasani and Stephen Casper and Tom Davidson and Raymond Douglas and David Duvenaud and Philip Fox and Usman Gohar and Rose Hadshar and Anson Ho and Tiancheng Hu and Cameron Jones and Sayash Kapoor and Atoosa Kasirzadeh and Sam Manning and Nestor Maslej and Vasilios Mavroudis and Conor McGlynn and Richard Moulange and Jessica Newman and Kwan Yee Ng and Patricia Paskov and Shalaleh Rismani and Girish Sastry and Elizabeth Seger and Scott Singer and Charlotte Stix and Lucia Velasco and Nicole Wheeler and Daron Acemoglu and Vincent Conitzer and Thomas G. Dietterich and Fredrik Heintz and Geoffrey Hinton and Nick Jennings and Susan Leavy and Teresa Ludermir and Vidushi Marda and Helen Margetts and John McDermid and Jane Munga and Arvind Narayanan and Alondra Nelson and Clara Neppel and Sarvapali D. Ramchurn and Stuart Russell and Marietje Schaake and Bernhard Schölkopf and Alvaro Soto and Lee Tiedrich and Gaël Varoquaux and Andrew Yao and Ya-Qin Zhang and Leandro Angelo Aguirre and Olubunmi Ajala and Fahad Albalawi and Noora AlMalek and Christian Busch and Jonathan Collas and André Carlos Ponce de Leon Ferreira de Carvalho and Amandeep Gill and Ahmet Halit Hatip and Juha Heikkilä and Chris Johnson and Gill Jolly and Ziv Katzir and Mary N. Kerema and Hiroaki Kitano and Antonio Krüger and Kyoung Mu Lee and José Ramón López Portillo and Aoife McLysaght and Oleksii Molchanovskyi and Andrea Monti and Mona Nemer and Nuria Oliver and Raquel Pezoa and Audrey Plonk and Balaraman Ravindran and Hammam Riza and Crystal Rugege and Haroon Sheikh and Denise Wong and Yi Zeng and Liming Zhu and Daniel Privitera and Sören Mindermann},
      year={2026},
      eprint={2602.21012},
      archivePrefix={arXiv},
      primaryClass={cs.CY},
      url={https://arxiv.org/abs/2602.21012}, 
}

@article{bickmore2018patient,
  title={Patient and consumer safety risks when using conversational assistants for medical information: an observational study of Siri, Alexa, and Google Assistant},
  author={Bickmore, Timothy W and Trinh, Ha and Olafsson, Stefan and O'Leary, Teresa K and Asadi, Reza and Rickles, Nathaniel M and Cruz, Ricardo},
  journal={Journal of medical Internet research},
  volume={20},
  number={9},
  pages={e11510},
  year={2018},
  publisher={JMIR Publications Inc., Toronto, Canada}
}

@misc{brodeur2026prospectiveclinicalfeasibilitystudy,
      title={A prospective clinical feasibility study of a conversational diagnostic AI in an ambulatory primary care clinic}, 
      author={Peter Brodeur and Jacob M. Koshy and Anil Palepu and Khaled Saab and Ava Homiar and Roma Ruparel and Charles Wu and Ryutaro Tanno and Joseph Xu and Amy Wang and David Stutz and Wei-Hung Weng and Hannah M. Ferrera and David Barrett and Lindsey Crowley and Jihyeon Lee and Spencer E. Rittner and Ellery Wulczyn and Selena K. Zhang and Elahe Vedadi and Christine G. Kohn and Kavita Kulkarni and Vinay Kadiyala and Sara Mahdavi and Wendy Du and Jessica M. Williams and David Feinbloom and Renee Wong and Tao Tu and Petar Sirkovic and Alessio Orlandi and Christopher Semturs and Yun Liu and Juraj Gottweis and Dale R. Webster and Joëlle Barral and Katherine Chou and Pushmeet Kohli and Avinatan Hassidim and Yossi Matias and James Manyika and Rob Fields and Jonathan X. Li and Marc L. Cohen and Vivek Natarajan and Mike Schaekermann and Alan Karthikesalingam and Adam Rodman},
      year={2026},
      eprint={2603.08448},
      archivePrefix={arXiv},
      primaryClass={cs.HC},
      url={https://arxiv.org/abs/2603.08448}, 
}

@article{brown-etal-1990-statistical,
    title = "A Statistical Approach to Machine Translation",
    author = "Brown, Peter F.  and
      Cocke, John  and
      Della Pietra, Stephen A.  and
      Della Pietra, Vincent J.  and
      Jelinek, Fredrick  and
      Lafferty, John D.  and
      Mercer, Robert L.  and
      Roossin, Paul S.",
    journal = "Computational Linguistics",
    volume = "16",
    number = "2",
    year = "1990",
    url = "https://aclanthology.org/J90-2002/",
    pages = "79--85"
}

@inproceedings{carenini-2000-task,
    title = "A Task-based Framework to Evaluate Evaluative Arguments",
    author = "Carenini, Giuseppe",
    editor = "Elhadad, Michael",
    booktitle = "{INLG}{'}2000 Proceedings of the First International Conference on Natural Language Generation",
    month = jun,
    year = "2000",
    address = "Mitzpe Ramon, Israel",
    publisher = "Association for Computational Linguistics",
    url = "https://aclanthology.org/W00-1402/",
    doi = "10.3115/1118253.1118256",
    pages = "9--16"
}

@inproceedings{cheng-mellish-2000-empirical,
    title = "An Empirical Analysis of Constructing Non-restrictive {NP} Modifiers to Express Semantic Relations",
    author = "Cheng, Hua  and
      Mellish, Chris",
    editor = "Elhadad, Michael",
    booktitle = "{INLG}{'}2000 Proceedings of the First International Conference on Natural Language Generation",
    month = jun,
    year = "2000",
    address = "Mitzpe Ramon, Israel",
    publisher = "Association for Computational Linguistics",
    url = "https://aclanthology.org/W00-1415/",
    doi = "10.3115/1118253.1118269",
    pages = "108--115"
}

@article{cheng2025benchmarking,
  title={Benchmarking is Broken--Don't Let AI be its Own Judge},
  author={Cheng, Zerui and Wohnig, Stella and Gupta, Ruchika and Alam, Samiul and Abdullahi, Tassallah and Ribeiro, Jo{\~a}o Alves and Nielsen-Garcia, Christian and Mir, Saif and Li, Siran and Orender, Jason and others},
  journal={arXiv preprint arXiv:2510.07575, presented at Neurips2025},
  year={2025}
}

@article{duggan:Jama,
    author = {Duggan, Matthew J. and Gervase, Julietta and Schoenbaum, Anna and Hanson, William and Howell, John T., III and Sheinberg, Michael and Johnson, Kevin B.},
    title = {Clinician Experiences With Ambient Scribe Technology to Assist With Documentation Burden and Efficiency},
    journal = {JAMA Network Open},
    volume = {8},
    number = {2},
    pages = {e2460637-e2460637},
    year = {2025},
    month = {02},
    issn = {2574-3805},
    doi = {10.1001/jamanetworkopen.2024.60637},
    url = {https://doi.org/10.1001/jamanetworkopen.2024.60637},
    eprint = {https://jamanetwork.com/journals/jamanetworkopen/articlepdf/2830383/duggan_2025_oi_241690_1739220983.37699.pdf},
}

@inproceedings{dusek-kasner-2020-evaluating,
    title = "Evaluating Semantic Accuracy of Data-to-Text Generation with Natural Language Inference",
    author = "Du{\v{s}}ek, Ond{\v{r}}ej  and
      Kasner, Zden{\v{e}}k",
    editor = "Davis, Brian  and
      Graham, Yvette  and
      Kelleher, John  and
      Sripada, Yaji",
    booktitle = "Proceedings of the 13th International Conference on Natural Language Generation",
    month = dec,
    year = "2020",
    address = "Dublin, Ireland",
    publisher = "Association for Computational Linguistics",
    url = "https://aclanthology.org/2020.inlg-1.19/",
    doi = "10.18653/v1/2020.inlg-1.19",
    pages = "131--137",
}

@article{gao-etal-2025-llm,
    title = "{LLM}-based {NLG} Evaluation: Current Status and Challenges",
    author = "Gao, Mingqi  and
      Hu, Xinyu  and
      Yin, Xunjian  and
      Ruan, Jie  and
      Pu, Xiao  and
      Wan, Xiaojun",
    journal = "Computational Linguistics",
    volume = "51",
    month = jun,
    year = "2025",
    address = "Cambridge, MA",
    publisher = "MIT Press",
    url = "https://aclanthology.org/2025.cl-2.9/",
    doi = "10.1162/coli_a_00561",
    pages = "661--687",
}

@inproceedings{gildea-jurafsky-2000-automatic,
    title = "Automatic Labeling of Semantic Roles",
    author = "Gildea, Daniel  and
      Jurafsky, Daniel",
    booktitle = "Proceedings of the 38th Annual Meeting of the Association for Computational Linguistics",
    month = oct,
    year = "2000",
    address = "Hong Kong",
    publisher = "Association for Computational Linguistics",
    url = "https://aclanthology.org/P00-1065/",
    doi = "10.3115/1075218.1075283",
    pages = "512--520"
}

@article{greenhaigh1997qualitative,
  title={Papers that go Beyond Numbers (Qualitative Research)'},
  author={Greenhaigh, T and Taylor, Rod},
  journal={British Medical Journal},
  volume={315},
  number={7110},
  pages={740--743},
  year={1997}
}

@book{guest2011applied,
  title={Applied thematic analysis},
  author={Guest, Greg and MacQueen, Kathleen M and Namey, Emily E},
  year={2011},
  publisher={sage publications}
}

@inproceedings{howcroft-etal-2020-twenty,
    title = "Twenty Years of Confusion in Human Evaluation: {NLG} Needs Evaluation Sheets and Standardised Definitions",
    author = "Howcroft, David M.  and
      Belz, Anya  and
      Clinciu, Miruna-Adriana  and
      Gkatzia, Dimitra  and
      Hasan, Sadid A.  and
      Mahamood, Saad  and
      Mille, Simon  and
      van Miltenburg, Emiel  and
      Santhanam, Sashank  and
      Rieser, Verena",
    editor = "Davis, Brian  and
      Graham, Yvette  and
      Kelleher, John  and
      Sripada, Yaji",
    booktitle = "Proceedings of the 13th International Conference on Natural Language Generation",
    month = dec,
    year = "2020",
    address = "Dublin, Ireland",
    publisher = "Association for Computational Linguistics",
    url = "https://aclanthology.org/2020.inlg-1.23/",
    doi = "10.18653/v1/2020.inlg-1.23",
    pages = "169--182",
}

@inproceedings{mani-etal-1999-tipster,
    title = "The {TIPSTER} {SUMMAC} Text Summarization Evaluation",
    author = "Mani, Inderjeet  and
      House, David  and
      Klein, Gary  and
      Hirschman, Lynette  and
      Firmin, Therese  and
      Sundheim, Beth",
    editor = "Thompson, Henry S.  and
      Lascarides, Alex",
    booktitle = "Ninth Conference of the {E}uropean Chapter of the Association for Computational Linguistics",
    month = jun,
    year = "1999",
    address = "Bergen, Norway",
    publisher = "Association for Computational Linguistics",
    url = "https://aclanthology.org/E99-1011/",
    pages = "77--85"
}

@inproceedings{mathur-etal-2020-tangled,
    title = "Tangled up in {BLEU}: Reevaluating the Evaluation of Automatic Machine Translation Evaluation Metrics",
    author = "Mathur, Nitika  and
      Baldwin, Timothy  and
      Cohn, Trevor",
    editor = "Jurafsky, Dan  and
      Chai, Joyce  and
      Schluter, Natalie  and
      Tetreault, Joel",
    booktitle = "Proceedings of the 58th Annual Meeting of the Association for Computational Linguistics",
    month = jul,
    year = "2020",
    address = "Online",
    publisher = "Association for Computational Linguistics",
    url = "https://aclanthology.org/2020.acl-main.448/",
    doi = "10.18653/v1/2020.acl-main.448",
    pages = "4984--4997",
}

@inproceedings{mccoy-etal-1990-using,
    title = "Using {T}ree {A}djoining {G}rammars Systemic Framework in the",
    author = "McCoy, Kathleen F.  and
      Vijay-Shanker, K.  and
      Yang, Gijoo",
    editor = "McKeown, Kathleen R.  and
      Moore, Johanna D.  and
      Nirenburg, Sergei",
    booktitle = "Proceedings of the Fifth International Workshop on Natural Language Generation",
    month = jun,
    year = "1990",
    address = "Linden Hall Conference Center, Dawson, Pennsylvania",
    publisher = "Association for Computational Linguistics",
    url = "https://aclanthology.org/W90-0101/"
}

@article{van-miltenburg-etal-2023-barriers,
    title = "Barriers and enabling factors for error analysis in {NLG} research",
    author = {van Miltenburg, Emiel  and
      Clinciu, Miruna  and
      Du{\v{s}}ek, Ond{\v{r}}ej  and
      Gkatzia, Dimitra  and
      Inglis, Stephanie  and
      Lepp{\"a}nen, Leo  and
      Mahamood, Saad  and
      Schoch, Stephanie  and
      Thomson, Craig  and
      Wen, Luou},
    editor = "Derczynski, Leon",
    journal = "Northern European Journal of Language Technology",
    volume = "9",
    year = "2023",
    address = {Link{\"o}ping, Sweden},
    publisher = {Link{\"o}ping University Electronic Press},
    url = "https://aclanthology.org/2023.nejlt-1.3/",
    doi = "10.3384/nejlt.2000-1533.2023.4529",
}

@inproceedings{minnen-etal-2000-robust,
    title = "Robust, applied morphological generation",
    author = "Minnen, Guido  and
      Carroll, John  and
      Pearce, Darren",
    editor = "Elhadad, Michael",
    booktitle = "{INLG}{'}2000 Proceedings of the First International Conference on Natural Language Generation",
    month = jun,
    year = "2000",
    address = "Mitzpe Ramon, Israel",
    publisher = "Association for Computational Linguistics",
    url = "https://aclanthology.org/W00-1427/",
    doi = "10.3115/1118253.1118281",
    pages = "201--208"
}

@inproceedings{murray-etal-2010-generating,
    title = "Generating and Validating Abstracts of Meeting Conversations: a User Study",
    author = "Murray, Gabriel  and
      Carenini, Giuseppe  and
      Ng, Raymond",
    editor = "Kelleher, John  and
      Namee, Brian Mac  and
      Sluis, Ielka van der",
    booktitle = "Proceedings of the 6th International Natural Language Generation Conference",
    month = jul,
    year = "2010",
    publisher = "Association for Computational Linguistics",
    url = "https://aclanthology.org/W10-4211/"
}

@misc{recent-frontier-models-are-reward-hacking,
    title = {Recent Frontier Models Are Reward Hacking},
    author = {Sydney Von Arx and Lawrence Chan and Elizabeth Barnes},
    howpublished = {\url{https://metr.org/blog/2025-06-05-recent-reward-hacking/}},
    year = {2025},
    month = {06},
}

@article{reddy2021evaluation,
  title={Evaluation framework to guide implementation of AI systems into healthcare settings},
  author={Reddy, Sandeep and Rogers, Wendy and Makinen, Ville-Petteri and Coiera, Enrico and Brown, Pieta and Wenzel, Markus and Weicken, Eva and Ansari, Saba and Mathur, Piyush and Casey, Aaron and others},
  journal={BMJ health \& care informatics},
  volume={28},
  number={1},
  pages={e100444},
  year={2021}
}

@article{reiter-belz-2009-investigation,
    title = "An Investigation into the Validity of Some Metrics for Automatically Evaluating Natural Language Generation Systems",
    author = "Reiter, Ehud  and
      Belz, Anja",
    journal = "Computational Linguistics",
    volume = "35",
    number = "4",
    month = dec,
    year = "2009",
    address = "Cambridge, MA",
    publisher = "MIT Press",
    url = "https://aclanthology.org/J09-4008/",
    doi = "10.1162/coli.2009.35.4.35405",
    pages = "529--558"
}

@inproceedings{reiter-etal-2001-using,
    title = "Using a Randomised Controlled Clinical Trial to Evaluate an {NLG} System",
    author = "Reiter, Ehud  and
      Robertson, Roma  and
      Lennox, A. Scott  and
      Osman, Liesl",
    booktitle = "Proceedings of the 39th Annual Meeting of the Association for Computational Linguistics",
    month = jul,
    year = "2001",
    address = "Toulouse, France",
    publisher = "Association for Computational Linguistics",
    url = "https://aclanthology.org/P01-1057/",
    doi = "10.3115/1073012.1073069",
    pages = "442--449"
}

@book{reiter2025,
author = {Ehud Reiter},
title = {Natural Language Generation},
year = 2025,
publisher = {Springer}
}

@article{reiter-2025-evaluate,
    title = "We Should Evaluate Real-World Impact",
    author = "Reiter, Ehud",
    journal = "Computational Linguistics",
    volume = "51",
    number = "4",
    month = dec,
    year = "2025",
    address = "Cambridge, MA",
    publisher = "MIT Press",
    url = "https://aclanthology.org/2025.cl-4.10/",
    doi = "10.1162/coli.a.18",
    pages = "1419--1431",
}

@book{waibel1990readings,
  title={Readings in speech recognition},
  author={Waibel, Alexander and Lee, Kai-Fu},
  year={1990},
  publisher={Morgan Kaufmann}
}

\end{document}